\PassOptionsToPackage{numbers}{natbib}
\documentclass[graybox, natbib]{svmult}

\usepackage{makeidx}         
\usepackage{graphicx}        
\usepackage{multicol}        
\usepackage[bottom, stable]{footmisc}

\usepackage{newtxtext}       %

\usepackage{booktabs}
\usepackage{longtable}
\usepackage{colortbl}
\usepackage{lmodern}
\usepackage{commath}
\usepackage{amsfonts}
\usepackage{xcolor}
\usepackage{comment}
\usepackage[hyphens]{url}

\newsavebox\CBox
\def\textBF#1{\sbox\CBox{#1}\resizebox{\wd\CBox}{\ht\CBox}{\textbf{#1}}}

\definecolor{lightblue}{rgb}{0.01, 0.6, 1.0}

\makeindex             

\begin{document}

\title*{Predicting Patient Outcomes with Graph Representation Learning}

\author{
    Emma Rocheteau\footnote[1]{Equal first authorship.}, 
    Catherine Tong\footnotemark[1], 
    Petar Veli\v{c}kovi\'{c}, 
    Nicholas Lane, 
    Pietro Li\`{o}
}

\authorrunning{Rocheteau, Tong \emph{et al.}}
\institute{
Emma Rocheteau \at University of Cambridge, \email{ecr38@cam.ac.uk}
\and Catherine Tong \at University of Oxford, \email{eu.tong@cs.ox.ac.uk}
\and Petar Veli\v{c}kovi\'{c} \at University of Cambridge
\and Nicholas Lane \at University of Cambridge, Samsung AI Center
\and Pietro Li\`{o} \at University of Cambridge
}

\maketitle

\abstract{Recent work on predicting patient outcomes in the Intensive Care Unit (ICU) has focused heavily on the physiological time series data, largely ignoring sparse data such as diagnoses and medications. When they are included, they are usually concatenated in the late stages of a model, which may struggle to learn from rarer disease patterns. Instead, we propose a strategy to exploit diagnoses as relational information by connecting similar patients in a graph. To this end, we propose LSTM-GNN for patient outcome prediction tasks: a hybrid model combining Long Short-Term Memory networks (LSTMs) for extracting temporal features and Graph Neural Networks (GNNs) for extracting the patient neighbourhood information. We demonstrate that LSTM-GNNs outperform the LSTM-only baseline on length of stay prediction tasks on the eICU database. More generally, our results indicate that exploiting information from neighbouring patient cases using graph neural networks is a promising research direction, yielding tangible returns in supervised learning performance on Electronic Health Records.}

\section{Introduction}
The past decade has seen growing interest in patient outcome prediction, particularly in the Intensive Care Unit (ICU). This is following the increased availability of Electronic Health Records (EHRs) and the drive to minimise preventable deaths through the use of early warning systems \citep{MOON2011150, doi:10.1513/AnnalsATS.201403-102OC}. Most prior works have focused on a small subset of features in the EHR \citep{Pencina2016} -- namely, the physiological time series data (especially following the publication of pre-processing pipelines e.g. \citet{harutyunyan}). This is problematic because the resulting models can miss clinically important information, leading to poorer clinical outcomes \citep{Rajkomar2018}.

\begin{figure}[h!]
	\centering
	\includegraphics[width=0.75\textwidth]{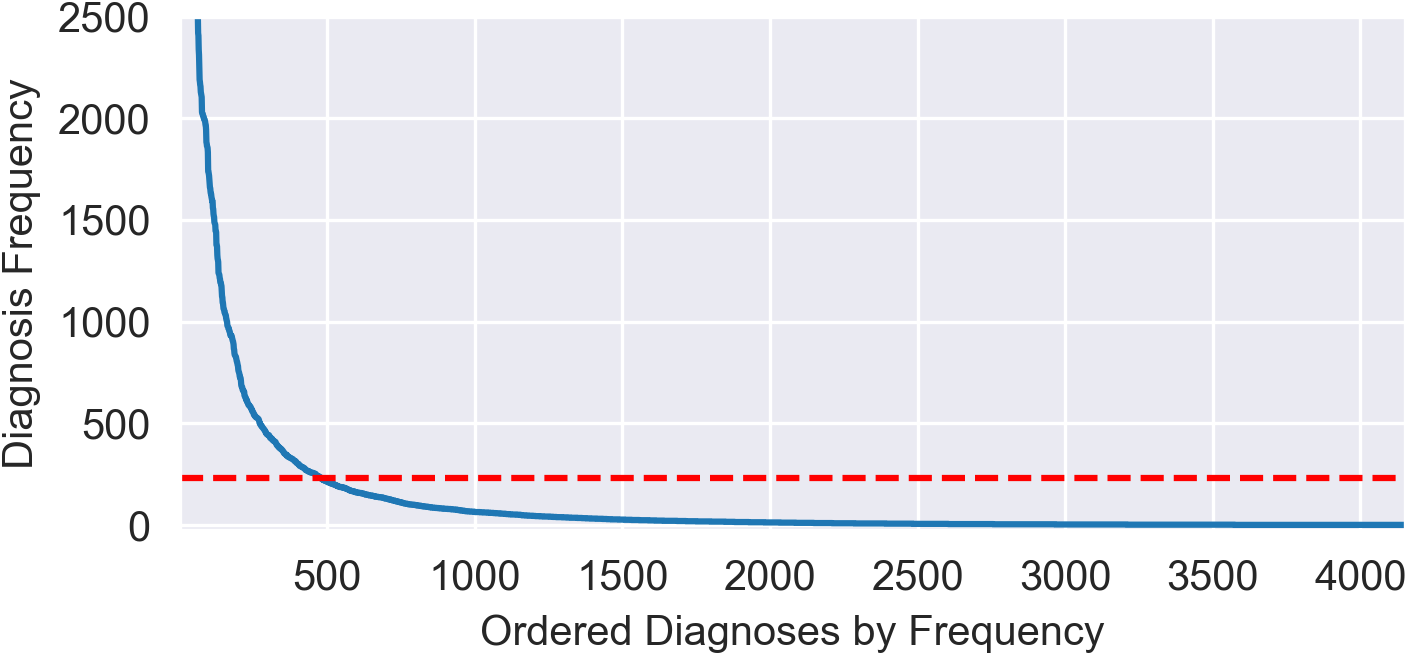}
	\caption{\small The distribution of diagnosis occurrence in our data is positively skewed. The mean number of samples per diagnosis is 229 (shown in red) which is not enough for most deep learning models to learn from. Note that the y axis has been truncated (the maximum value is in fact 79,778).}
    \label{diagdist}
\end{figure}

Among the frequently overlooked (but very informative) data are the diagnoses, medications and surgical procedures. They are difficult to use for two reasons:
\begin{enumerate}
    \item The large number of features makes distinguishing relevant comorbidity patterns combinatorially difficult\footnote{The average patient in eICU has 9 recorded diagnoses, but there are 4,172 distinct diagnoses in our cohort.}.
    \item The model does not have enough data on rare diseases.
\end{enumerate}
The common approach has been to throw away the long tail of the distribution (shown in Figure~\ref{diagdist}) during pre-processing and concatenate the remaining features (often via an encoder network) to the main part of the model at a late stage. Unfortunately this approach is always a compromise between the difficulty of the modelling task and the amount of valuable data that is thrown away.

We want to improve on this approach toward sparse information in the EHR, taking diagnoses as an example. In the design of our model we take inspiration from clinicians, who tend to rely on their past experience of treating similar patients when making clinical judgements. We capture this similarity concept by constructing a \emph{patient graph} where the nodes are patients and the edges express relatedness in diagnoses. We exploit this information with a hybrid architecture consisting of a Long Short-Term Memory (LSTM) \citep{hochreiter1997long} network composed with a Graph Neural Network (GNN) \citep{gori2005new,scarselli2009graph} to predict in-hospital mortality (IHM) and length of ICU stay (LOS) using data from the first 24 hours of the ICU stay. This represents a novel application of GNNs in healthcare. While we focus on diagnosis information, our method can easily be extended to other sparse medical data such as shared medications. Our code can be found at \url{https://github.com/EmmaRocheteau/eICU-GNN-LSTM}.

\section{Related Work}

Our work is motivated by the following areas of related work:

\emph{Graph neural networks} (GNNs) are a subclass of neural networks which operate on graph-structured data as input. The general principle is to apply a transformation function to each node representation in the graph, before aggregating information between neighbouring nodes. Different GNNs vary in their node transformation and neighbourhood aggregation functions \citep{xu2018how}. In our work, we select four popular GNNs to model the similarity relationships between patients: Graph Convolutional Networks (GCN)~\citep{kipf2016semi}, Graph Attention Networks (GAT)~\citep{velivckovic2017graph}, GraphSAGE~\citep{hamilton2017inductive}, and Message Passing Neural Networks (MPNN)~\citep{gilmer_mpnn}.

\emph{Recurrent Neural Networks} (particularly LSTMs) have so far been the most popular model for patient outcome prediction from time series, and they have achieved state-of-the-art results on the MIMIC and eICU datasets \citep{harutyunyan,Rajkomar2018,sheikhalishahi2019benchmarking}. We therefore select an LSTM (very similar to that used in \citet{harutyunyan}) to model the time series component.

\emph{Combination Models} combining sequential modelling with GNNs has been explored by several works outside of the healthcare domain in recent years \citep{Goyal_2020,li2018diffusion,pareja2019evolvegcn,spatiotemporal}. However, we stress that our graph is static (a reasonable assumption since patient diagnoses do not vary much during an ICU stay), which calls for a simpler modelling approach than these works.

\emph{Graphical representation of clinical data} is a young and exciting research domain \citep{Schrodt2020} that has so far focused on injecting medical knowledge in the form of knowledge graphs, or structuring the EHR itself as a graph, as in \citet{choi2020} (note that these applications do not employ any \emph{inter-patient} data sharing). The only example of a patient graph in the literature is \citet{malone2018learning} who solve the task of missing data imputation using embedding propagation \citep{duran2017learning}. This is done as a separate step (i.e. their approach is not end-to-end) before using logistic regression and ridge regression as downstream prediction models.

\section{Methods}

\begin{figure}[t]
	\centering
	\includegraphics[width=0.685\textwidth]{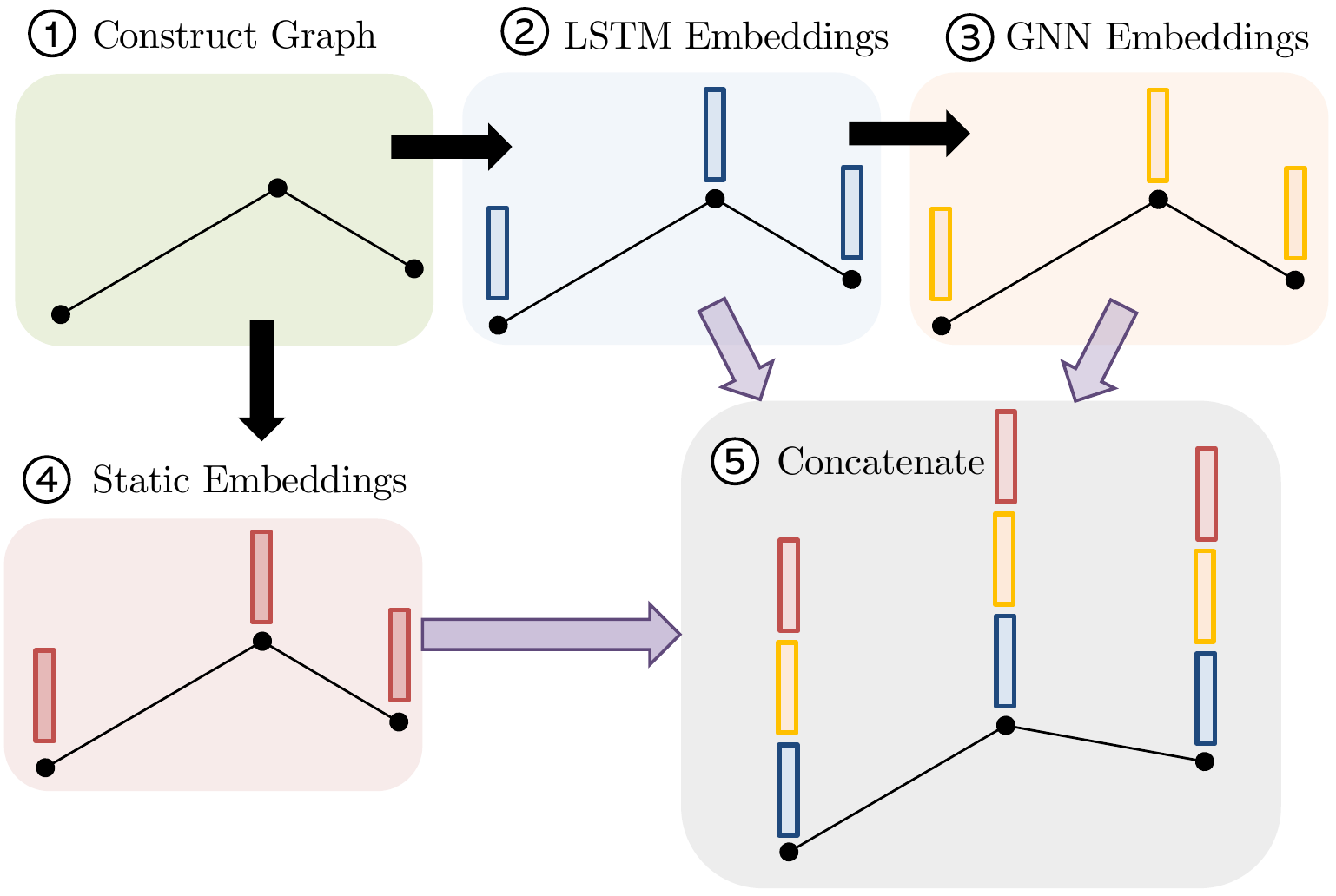}
	\caption{\small Approach Overview. First, we construct a patient graph, which becomes the input to an end-to-end LSTM-GNN model. Through the LSTM-GNN, each node's temporal features are encoded by a temporal encoder followed by a graph encoder, and the static features are encoded separately. Finally, these are concatenated and passed to a fully-connected layer for prediction.}
	\label{fig:overall}
\end{figure}

Figure~\ref{fig:overall} gives an overview of our approach. We start with a set of static and temporal features for each patient. Our first step (top-left in Figure~\ref{fig:overall}) is to define a patient graph construction, $\mathcal{G}$, where related patients (nodes) are connected by edges. Next, we pass the patient graph as input to the LSTM-GNN, which is trained end-to-end. The LSTM-GNN produces three types of embeddings: the LSTM, GNN and static embeddings, which are concatenated and passed to a fully-connected layer to obtain per-node predictions. In the following, we provide the technical details of both the graph construction and LSTM-GNN training procedures.

\vspace{0.5em}

\runinhead{Diagnosis Graph Construction}
We start by assigning a pairwise similarity score between all patients. First, we transform the diagnoses into a multi-hot vector for each patient, resulting in a diagnosis matrix $\mathcal{D} \in \mathbb{R}^{N \times m}$ where $m$ is the number of unique diagnoses and $N$ is the number of patients. The similarity score $\mathcal{M}_{ij}$ between nodes $i$ and $j$ is defined as
\begin{equation} 
\label{eq:scoring}
\mathcal{M}_{ij} = a \overbrace{\sum^m_{\mu=1}\left(\mathcal{D}_{i\mu}\mathcal{D}_{j \mu}(d_\mu^{-1} + c)\right)}^{\text{Shared Diagnoses}} - \overbrace{\sum^m_{\mu=1}\left(\mathcal{D}_{i\mu} + \mathcal{D}_{j \mu}\right)}^{\text{All Diagnoses}}
\end{equation}
where $d_{\mu}$ is the occurrence of diagnosis $\mu$, and $a$ and $c$ are tunable constants. The first term positively rewards shared diagnoses. Note that the $d_\mu^{-1}$ term incorporates the idea that two patients sharing a rare diagnosis is more significant than a common one. The second term penalises the total number of diagnoses -- this is to prevent patients with many diagnoses becoming `hubs' of high connectivity, attracting imprecise matches with several non-shared diagnoses. We examine $\mathcal{M}$ under a $k$-Nearest Neighbour ($k$-NN) scheme to establish $k$ edges per node. The parameters $a$, $c$ and $k$ were treated as hyperparameters ($c = 0.001$, $a=5$ and $k=3$ in the final model).

We experimented with alternative graph construction methods, such as applying a score threshold for edges (more akin to \citet{malone2018learning}), or using BERT ~\citep{Devlin2019BERTPO} to encode diagnosis texts prior to similarity computation. Empirically we have found the presented method to work best through manual inspection of the resultant graph and preliminary experimentation. 

\vspace{0.5em}

\runinhead{LSTM-GNNs} \label{sec:model}

\begin{figure}[ht]
	\centering
	\includegraphics[width=0.97\textwidth]{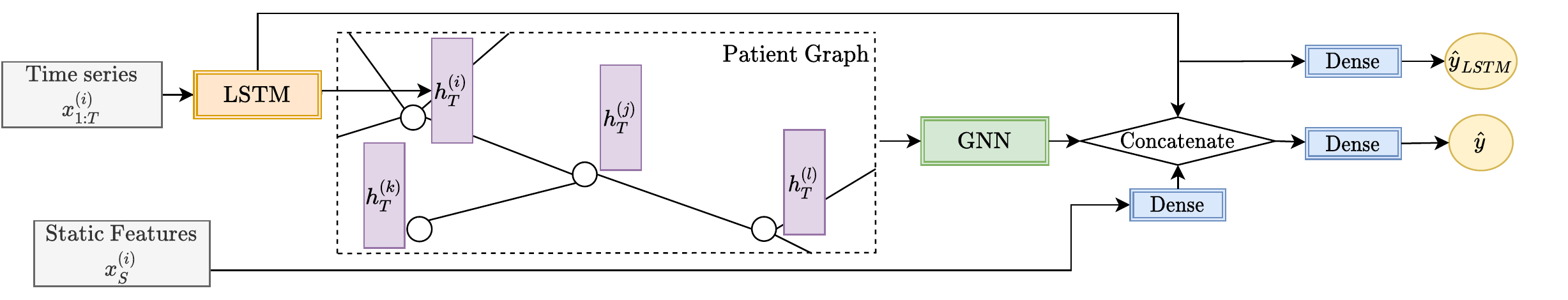}
	\caption{\small Our LSTM-GNN architecture. The LSTM extracts temporal features from time series data. These patient-level features are then propagated within local neighbourhoods by the GNN. We concatenate hidden vectors from the temporal, graph and the static features to make the prediction.}
	\label{fig:model}
\end{figure}

Having constructed the patient graph, we frame the patient outcome prediction problem as a node prediction task. We use LSTM-GNN, a hybrid model consisting of temporal and graph encoding components (summarised in Figure~\ref{fig:model}). We assume the input of LSTM-GNN to be a patient graph $\mathcal{G}$, with each node $i$ having time series $x^{(i)}_{1:T}$ and static features $x^{(i)}_S$ (this includes diagnoses and other variables e.g. age and gender). We describe a forward pass through the network as follows.

The time series $x^{(i)}_{1:T}$ are first passed through a bi-directional LSTM, which outputs a sequence of hidden state vectors $h^{(i)}_{1:T}$ in the forward and reverse directions. The vectors corresponding to the last timestep are concatenated to produce $h^{(i)}_T$, a temporal embedding per node.

Next, the GNN component propagates each node's temporal embedding within its neighbourhood. This function varies between GNNs, however, the aim is always to apply a local smoothing. That is, the features $h^{(i)}_T$ are re-weighted with the feature vectors in its local neighbourhood to produce the new node representation $h^{(i)}_N$.

Meanwhile, we pass the static input $x^{(i)}_S$ through a fully-connected layer to compute $h^{(i)}_S$, before concatenating our learnt representations together, $h^{(i)} = (h^{(i)}_T || h^{(i)}_N || h^{(i)}_S)$. Finally, $h^{(i)}$ is passed through a fully-connected layer to obtain a prediction $\hat{y}^{(i)}$. 

We train the LSTM-GNN in an end-to-end and scalable fashion. To allow for mini-batch training of the LSTM, we adopt a neighbourhood sampling procedure proposed by \citet{hamilton2017inductive}. With each iteration, we uniformly sample a fixed-size set of neighbours in the graph, which fixes computation and time costs per iteration, thus making the LSTM-GNN scalable to large patient graphs. We take an inductive learning approach. During training, we only sample nodes and their neighbourhood from the training set. During testing, we sample neighbours from the entire dataset but we only evaluate performance on the test nodes.

We noticed that the LSTM performance can degrade with the addition of a GNN. To encourage learning from both components, we define the loss function as

\begin{equation}
    \mathcal{L} = \mathcal{L}_{\text{LSTM-GNN}} + \alpha \mathcal{L}_{\text{LSTM}}
\end{equation}
where $\mathcal{L}_{\text{LSTM-GNN}}$ is the loss on the full model prediction $\hat{y}$, $\mathcal{L}_{\text{LSTM}}$ is the loss on the prediction made by the LSTM component $\hat{y}_{\text{LSTM}}$ (computed by passing $h_T$ through a distinct fully-connected layer), and $\alpha$ is treated as a hyperparameter. For IHM, the loss function is binary crossentropy, whereas for LOS it is the squared logarithmic error (found to mitigate for positive skew in \citet{rocheteau2020temporal}). 

\section{Data}
We use the eICU Collaborative Research Database \citep{Pollard2018}, a multi-centre dataset collated from 208 hospitals in the United States. We selected flat features, time series and diagnoses from 89,123 adult patients (\textgreater 18 years) with an ICU LOS of at least 24 hours and one recorded observation. If the patient had multiple admissions we selected one at random. The dataset was divided at the patient level such that 70\%, 15\% and 15\% were used for training, validation and testing respectively.

\runinhead{Static Features}
We initially extracted 20 non-time varying features (shown in Table~\ref{tab:static}). Discrete and continuous variables were scaled to the interval [-1, 1], using the 5th and 95th percentiles as the boundaries, and absolute cut offs were placed at [-4, 4]. This was to protect against large or erroneous inputs, while avoiding assumptions about the variable distributions. Binary variables were coded as 1 and 0. Categorical variables were converted to one-hot encodings.

\subruninhead{Diagnoses}
Like many EHRs, diagnosis coding in eICU is hierarchical. At the lowest level they are very specific e.g.\ ``neurologic $|$ disorders of vasculature $|$ stroke $|$ hemorrhagic stroke $|$ subarachnoid hemorrhage $|$ with vasospasm''. To maintain the hierarchical structure, we assigned separate features to each class level and used multi-hot encodings with each position referring to a particular diagnosis. This produces a vector of size 4,436 with an average sparsity of 99.5\%. We include diagnoses with a prevalence greater than 0.5\%. If a disease does not make this threshold, it is still included via any parent classes that do qualify (e.g.\ in the above example we retain everything up to ``subarachnoid hemorrhage''). We only included diagnoses that were recorded before the 24th hour in the ICU.

\runinhead{Time Series}
\label{timeseriespreproc}
For each admission, 18 time-varying features (Table~\ref{tab:timeseries}) were extracted from each hour of the stay, and up to 24 hours before. The variables were processed in the same manner as the static features. In general, the sampling was irregular, so the data was re-sampled according to one hour intervals and forward-filled.

\begin{table}
\parbox{.465\linewidth}{
    \caption{\small Non-time varying features. Age \textgreater 89, Null Height and Null Weight were added as indicator variables to indicate when the age was more than 89 but has been capped, and when the height or weight were missing and have been imputed with the mean value.}
    \label{tab:static}
    \centering
    \scriptsize
    \begin{tabular}{lll}
        \toprule
        \textbf{Feature} & \textbf{Type} & \textbf{Source Table} \\
        \midrule
        Gender & Binary & \textit{patient} \\
        Age & Discrete & \textit{patient} \\
        Hour of Admission & Discrete & \textit{patient} \\
        Height & Continuous & \textit{patient} \\
        Weight & Continuous & \textit{patient} \\
        Ethnicity & Categorical & \textit{patient} \\
        Unit Type & Categorical & \textit{patient} \\
        Unit Admit Source & Categorical & \textit{patient} \\
        Unit Stay Type & Categorical & \textit{patient} \\
        Physician Speciality & Categorical & \textit{apachepatientresult} \\
        Eyes & Discrete & \textit{apacheapsvar} \\
        Motor & Discrete & \textit{apacheapsvar} \\
        Verbal & Discrete & \textit{apacheapsvar} \\
        Meds & Discrete & \textit{apacheapsvar} \\
        Intubated & Binary & \textit{apacheapsvar} \\
        Ventilated & Binary & \textit{apacheapsvar} \\
        Dialysis & Binary & \textit{apacheapsvar} \\
        Age \textgreater 89 & Binary & \\
        Null Height & Binary & \\
        Null Weight & Binary & \\
        \bottomrule
    \end{tabular}
}
\hfill
\parbox{.465\linewidth}{
    \caption{\small Time series features. `Time in the ICU' and `Time of day' were not part of the tables in eICU but were added later as helpful indicators to the model.}
    \label{tab:timeseries}
    \centering
    \scriptsize
    \begin{tabular}{lll}
        \toprule
        \textbf{Feature} & \textbf{Source Table} \\
        \midrule
        Bedside Glucose & \textit{lab} \\
        FiO2 & \textit{respiatorycharting} \\
        SaO2 & \textit{respiratorycharting} \\
        Non-Invasive Diastolic & \textit{vitalaperiodic} \\
        Non-Invasive Mean & \textit{vitalaperiodic} \\
        Non-Invasive Systolic & \textit{vitalaperiodic} \\
        CVP & \textit{vitalperiodic} \\
        Heart Rate & \textit{vitalperiodic} \\
        Respiration & \textit{vitalperiodic} \\
        st1 & \textit{vitalperiodic} \\
        st2 & \textit{vitalperiodic} \\
        st3 & \textit{vitalperiodic} \\
        Systemic Diastolic & \textit{vitalperiodic} \\
        Systemic Mean & \textit{vitalperiodic} \\
        Systemic Systolic & \textit{vitalperiodic} \\
        Temperature & \textit{vitalperiodic} \\
        Time in the ICU & \\
        Time of day & \\
        \bottomrule
    \end{tabular}
}
\end{table}

\section{Results}
\begin{trailer}{Evaluation Metrics}
\small
\emph{In-Hospital Mortality}:
\begin{itemize}
    \item Area under the receiver operating characteristic curve (AUROC)
    \item Area under the precision recall curve (AUPRC)
\end{itemize}
\emph{Length of Stay}:
\begin{itemize}
    \item Mean absolute deviation (MAD)
    \item Mean absolute percentage error (MAPE)
    \item Mean squared error (MSE)
    \item Mean squared log error (MSLE)
    \item Coefficient of determination ($R^2$)
    \item Cohen's linear weighted Kappa Score \citep{doi:10.1177/001316446002000104} (Kappa) (as applied in \citet{harutyunyan}).
\end{itemize}
    For AUROC, AUPRC, $R^2$ and Kappa, higher is better, whereas for MAD, MAPE, MSE and MSLE lower is better.
\end{trailer}

\runinhead{LSTM-GNN Performance}

\begin{table*}[hbt!]
  \caption{\small Performance of various LSTM-GNN models. We compare these models to an LSTM baseline (with and without* diagnosis concatenation). The error margins are 95\% confidence intervals (CIs) from 15 independent training runs. The best results are highlighted in blue. If a result is statistically different from the \emph{LSTM} (with diagnoses) on a two-tailed t-test (p < 0.05), then it is indicated with $\ddag$ or $\dag$ to show better or worse performance respectively.}
  \label{tab:results}
  \centering
  \scriptsize
  \begin{tabular}{p{1.95cm}p{1.2cm}p{1.17cm}p{1cm}p{0.87cm}p{0.87cm}p{1.2cm}p{1.2cm}p{1.2cm}}
    \toprule
        & \multicolumn{2}{c}{\scriptsize{\textBF{In-Hospital Mortality}}} & \multicolumn{6}{c}{\scriptsize{\textBF{Length of Stay}}} \\
        \scriptsize{\textBF{Model}} & \scriptsize{\textBF{AUROC}} & \scriptsize{\textBF{AUPRC}} & \scriptsize{\textBF{MAD}} & \scriptsize{\textBF{MAPE}} & \scriptsize{\textBF{MSE}} & \scriptsize{\textBF{MSLE}} & \scriptsize{\boldmath{$R^2$}} & \scriptsize{\textBF{Kappa}} \\
    \midrule
        \scriptsize{LSTM* (no diag.)} & {\fontsize{5.5}{5}\selectfont 0.837$\pm$0.001}$^\dag$ & {\fontsize{5.5}{5}\selectfont 0.390$\pm$0.004}$^\dag$ & {\fontsize{5.5}{5}\selectfont 1.97$\pm$0.01}$^\dag$ & {\fontsize{5.5}{5}\selectfont \textBF{\textcolor{blue}{49.4$\pm$0.6}}} & {\fontsize{5.5}{5}\selectfont 17.6$\pm$0.2}$^\dag$ & {\fontsize{5.5}{5}\selectfont 0.398$\pm$0.004}$^\dag$ & {\fontsize{5.5}{5}\selectfont 0.089$\pm$0.008}$^\dag$ & {\fontsize{5.5}{5}\selectfont 0.224$\pm$0.006}$^\dag$ \\
        \scriptsize{LSTM} & {\fontsize{5.5}{5}\selectfont \textBF{\textcolor{blue}{0.858$\pm$0.001}}} & {\fontsize{5.5}{5}\selectfont 0.429$\pm$0.002} & {\fontsize{5.5}{5}\selectfont 1.95$\pm$0.01} & {\fontsize{5.5}{5}\selectfont 49.8$\pm$0.9} & {\fontsize{5.5}{5}\selectfont 17.0$\pm$0.1} & {\fontsize{5.5}{5}\selectfont 0.382$\pm$0.001} & {\fontsize{5.5}{5}\selectfont 0.118$\pm$0.003} & {\fontsize{5.5}{5}\selectfont 0.245$\pm$0.006} \\
        \scriptsize{LSTM-SAGE} & {\fontsize{5.5}{5}\selectfont 0.851$\pm$0.003}$^\dag$ & {\fontsize{5.5}{5}\selectfont 0.426$\pm$0.010} & {\fontsize{5.5}{5}\selectfont 1.87$\pm$0.00}$^\ddag$ & {\fontsize{5.5}{5}\selectfont 50.9$\pm$0.5}$^\dag$ & {\fontsize{5.5}{5}\selectfont 14.8$\pm$0.1}$^\ddag$ & {\fontsize{5.5}{5}\selectfont 0.377$\pm$0.002}$^\ddag$ & {\fontsize{5.5}{5}\selectfont 0.119$\pm$0.005} & {\fontsize{5.5}{5}\selectfont 0.237$\pm$0.006} \\
        \scriptsize{LSTM-GAT} & {\fontsize{5.5}{5}\selectfont 0.854$\pm$0.001}$^\dag$ & {\fontsize{5.5}{5}\selectfont 0.427$\pm$0.003} & {\fontsize{5.5}{5}\selectfont \textBF{\textcolor{blue}{1.86$\pm$0.00}}}$^\ddag$ & {\fontsize{5.5}{5}\selectfont 49.7$\pm$0.3} & {\fontsize{5.5}{5}\selectfont 14.6$\pm$0.1}$^\ddag$ & {\fontsize{5.5}{5}\selectfont 0.371$\pm$0.001}$^\ddag$ & {\fontsize{5.5}{5}\selectfont 0.129$\pm$0.004}$^\ddag$ & {\fontsize{5.5}{5}\selectfont 0.258$\pm$0.004}$^\ddag$ \\
        \scriptsize{LSTM-MPNN} & {\fontsize{5.5}{5}\selectfont 0.852$\pm$0.001}$^\dag$ & {\fontsize{5.5}{5}\selectfont \textBF{\textcolor{blue}{0.433$\pm$0.004}}} & {\fontsize{5.5}{5}\selectfont \textBF{\textcolor{blue}{1.86$\pm$0.01}}}$^\ddag$ & {\fontsize{5.5}{5}\selectfont 50.5$\pm$1.3} & {\fontsize{5.5}{5}\selectfont \textBF{\textcolor{blue}{14.5$\pm$0.1}}}$^\ddag$ & {\fontsize{5.5}{5}\selectfont \textBF{\textcolor{blue}{0.369$\pm$0.001}}}$^\ddag$ & {\fontsize{5.5}{5}\selectfont \textBF{\textcolor{blue}{0.136$\pm$0.007}}}$^\ddag$ & {\fontsize{5.5}{5}\selectfont \textBF{\textcolor{blue}{0.261$\pm$0.005}}}$^\ddag$ \\
    \arrayrulecolor{black}\bottomrule
    \end{tabular}
\end{table*}

In Table~\ref{tab:results}, we compare the LSTM-GNN performance to LSTM baselines \citep{harutyunyan, sheikhalishahi2019benchmarking}. The first baseline (LSTM*) does not take any diagnoses as input\footnote{We use $*$ to denote all models which exclude diagnoses from input $x_S$.}, whereas the second baseline (LSTM) processes diagnoses according to the commonly applied encoder concatenation approach. LSTM significantly outperforms LSTM*, confirming that diagnoses add predictive value to both tasks.

All of the LSTM-GNN models demonstrate significant performance gains compared to both LSTM baselines on the \emph{LOS} task. LSTM-MPNN in particular demonstrates impressive performance\footnote{Note that the LSTM-MPNN model is the most expressive of the GNNs evaluated, as it has the capacity to model edge features (similarity scores from Equation~\ref{eq:scoring}) while other GNNs do not.}, surpassing LSTM by $3-15\%$ on all LOS metrics except MAPE. Additional investigation revealed that the error reduction in LSTM-GNN models corresponds to long LOSs, which explains the disproportionate reduction in MSE and not MAPE. On IHM, the LSTM-GNN models tend to show a small (but statistically insignificant) increase in AUPRC, but a reduction in AUROC.

\runinhead{Ablation Studies}

\begin{table}[ht!]
  \caption{\small Ablation Studies. (a) shows the performance of various LSTM-GNN* models (without diagnoses). The t-tests are performed with respect to LSTM*. (b) shows the results when GraphSAGE and GAT operate without an LSTM i.e. we provide the raw time series as input to the GNN. They are compared to their respective LSTM-GNN models.}
  \label{tab:ablation}
  \centering
  \scriptsize
  \begin{minipage}{0.04\textwidth}
  \centering
  \vspace{3.5em}
  (a)
  \end{minipage}
  \begin{minipage}{0.95\textwidth}
  \centering
  \begin{tabular}{p{1.6cm}p{1.2cm}p{1.17cm}p{1cm}p{0.87cm}p{0.87cm}p{1.2cm}p{1.2cm}p{1.2cm}}
    \toprule
        & \multicolumn{2}{c}{\scriptsize{\textBF{In-Hospital Mortality}}} & \multicolumn{6}{c}{\scriptsize{\textBF{Length of Stay}}} \\
        \scriptsize{\textBF{Model}} & \scriptsize{\textBF{AUROC}} & \scriptsize{\textBF{AUPRC}} & \scriptsize{\textBF{MAD}} & \scriptsize{\textBF{MAPE}} & \scriptsize{\textBF{MSE}} & \scriptsize{\textBF{MSLE}} & \scriptsize{\boldmath{$R^2$}} & \scriptsize{\textBF{Kappa}} \\
    \midrule
        \scriptsize{LSTM*} & {\fontsize{5.5}{5}\selectfont 0.837$\pm$0.001} & {\fontsize{5.5}{5}\selectfont 0.390$\pm$0.004} & {\fontsize{5.5}{5}\selectfont 1.97$\pm$0.01} & {\fontsize{5.5}{5}\selectfont \textBF{\textcolor{blue}{49.4$\pm$0.6}}} & {\fontsize{5.5}{5}\selectfont 17.6$\pm$0.2} & {\fontsize{5.5}{5}\selectfont 0.398$\pm$0.004} & {\fontsize{5.5}{5}\selectfont 0.089$\pm$0.008} & {\fontsize{5.5}{5}\selectfont 0.224$\pm$0.006} \\
        \scriptsize{LSTM-SAGE*} & {\fontsize{5.5}{5}\selectfont \textBF{\textcolor{blue}{0.840$\pm$0.001}}}$^\ddag$ & {\fontsize{5.5}{5}\selectfont \textBF{\textcolor{blue}{0.397$\pm$0.006}}}$^\ddag$ & {\fontsize{5.5}{5}\selectfont 1.88$\pm$0.01}$^\ddag$ & {\fontsize{5.5}{5}\selectfont 50.7$\pm$1.2} & {\fontsize{5.5}{5}\selectfont 14.9$\pm$0.1}$^\ddag$ & {\fontsize{5.5}{5}\selectfont 0.380$\pm$0.001}$^\ddag$ & {\fontsize{5.5}{5}\selectfont 0.117$\pm$0.006}$^\ddag$ & {\fontsize{5.5}{5}\selectfont 0.240$\pm$0.006}$^\ddag$ \\
        \scriptsize{LSTM-GAT*} & {\fontsize{5.5}{5}\selectfont 0.838$\pm$0.002} & {\fontsize{5.5}{5}\selectfont 0.384$\pm$0.008} & {\fontsize{5.5}{5}\selectfont 1.88$\pm$0.01}$^\ddag$ & {\fontsize{5.5}{5}\selectfont 50.1$\pm$1.4} & {\fontsize{5.5}{5}\selectfont 15.0$\pm$0.1}$^\ddag$ & {\fontsize{5.5}{5}\selectfont 0.383$\pm$0.003}$^\ddag$ & {\fontsize{5.5}{5}\selectfont 0.108$\pm$0.009}$^\ddag$ & {\fontsize{5.5}{5}\selectfont 0.234$\pm$0.005}$^\ddag$ \\
        \scriptsize{LSTM-MPNN*} & {\fontsize{5.5}{5}\selectfont 0.836$\pm$0.001} & {\fontsize{5.5}{5}\selectfont 0.392$\pm$0.003} & {\fontsize{5.5}{5}\selectfont \textBF{\textcolor{blue}{1.87$\pm$0.01}}}$^\ddag$ & {\fontsize{5.5}{5}\selectfont 50.8$\pm$1.7} & {\fontsize{5.5}{5}\selectfont \textBF{\textcolor{blue}{14.7$\pm$0.2}}}$^\ddag$ & {\fontsize{5.5}{5}\selectfont \textBF{\textcolor{blue}{0.377$\pm$0.004}}}$^\ddag$ & {\fontsize{5.5}{5}\selectfont \textBF{\textcolor{blue}{0.128$\pm$0.011}}}$^\ddag$ & {\fontsize{5.5}{5}\selectfont \textBF{\textcolor{blue}{0.255$\pm$0.008}}}$^\ddag$
    \end{tabular}
    \end{minipage}
    \begin{minipage}{0.04\textwidth}
    \label{tab:extra}
    \centering
    (b)
    \end{minipage}
    \begin{minipage}{0.95\textwidth}
    \centering
  \begin{tabular}{p{1.6cm}p{1.2cm}p{1.17cm}p{1cm}p{0.87cm}p{0.87cm}p{1.2cm}p{1.2cm}p{1.2cm}}
    \arrayrulecolor{black}\specialrule{0.7pt}{2pt}{3pt}
        \scriptsize{SAGE} & {\fontsize{5.5}{5}\selectfont \textBF{\textcolor{blue}{0.853$\pm$0.001}}} & {\fontsize{5.5}{5}\selectfont 0.406$\pm$0.003}$^\dag$ & {\fontsize{5.5}{5}\selectfont 1.96$\pm$0.00}$^\dag$ & {\fontsize{5.5}{5}\selectfont \textBF{\textcolor{blue}{50.7$\pm$0.9}}} & {\fontsize{5.5}{5}\selectfont 17.1$\pm$0.1}$^\dag$ & {\fontsize{5.5}{5}\selectfont 0.389$\pm$0.001}$^\dag$ & {\fontsize{5.5}{5}\selectfont 0.113$\pm$0.006} & {\fontsize{5.5}{5}\selectfont \textBF{\textcolor{blue}{0.239$\pm$0.004}}} \\
        \scriptsize{LSTM-SAGE} & {\fontsize{5.5}{5}\selectfont 0.851$\pm$0.003} & {\fontsize{5.5}{5}\selectfont \textBF{\textcolor{blue}{0.426$\pm$0.010}}} & {\fontsize{5.5}{5}\selectfont \textBF{\textcolor{blue}{1.87$\pm$0.00}}} & {\fontsize{5.5}{5}\selectfont 50.9$\pm$0.5} & {\fontsize{5.5}{5}\selectfont \textBF{\textcolor{blue}{14.8$\pm$0.1}}} & {\fontsize{5.5}{5}\selectfont \textBF{\textcolor{blue}{0.377$\pm$0.002}}} & {\fontsize{5.5}{5}\selectfont \textBF{\textcolor{blue}{0.119$\pm$0.005}}} & {\fontsize{5.5}{5}\selectfont 0.237$\pm$0.006} \\
        \arrayrulecolor{gray}\specialrule{0.2pt}{1.5pt}{1.5pt}
        \scriptsize{GAT} & {\fontsize{5.5}{5}\selectfont 0.833$\pm$0.001}$^\dag$ & {\fontsize{5.5}{5}\selectfont 0.357$\pm$0.003}$^\dag$ & {\fontsize{5.5}{5}\selectfont 2.02$\pm$0.01}$^\dag$ & {\fontsize{5.5}{5}\selectfont 52.2$\pm$1.0}$^\dag$ & {\fontsize{5.5}{5}\selectfont 18.0$\pm$0.1}$^\dag$ & {\fontsize{5.5}{5}\selectfont 0.423$\pm$0.003}$^\dag$ & {\fontsize{5.5}{5}\selectfont 0.066$\pm$0.006}$^\dag$ & {\fontsize{5.5}{5}\selectfont 0.186$\pm$0.006}$^\dag$ \\
        \scriptsize{LSTM-GAT} & {\fontsize{5.5}{5}\selectfont \textBF{\textcolor{blue}{0.854$\pm$0.001}}} & {\fontsize{5.5}{5}\selectfont \textBF{\textcolor{blue}{0.427$\pm$0.003}}} & {\fontsize{5.5}{5}\selectfont \textBF{\textcolor{blue}{1.86$\pm$0.00}}} & {\fontsize{5.5}{5}\selectfont \textBF{\textcolor{blue}{49.7$\pm$0.3}}} & {\fontsize{5.5}{5}\selectfont \textBF{\textcolor{blue}{14.6$\pm$0.1}}} & {\fontsize{5.5}{5}\selectfont \textBF{\textcolor{blue}{0.371$\pm$0.001}}} & {\fontsize{5.5}{5}\selectfont \textBF{\textcolor{blue}{0.129$\pm$0.004}}} & {\fontsize{5.5}{5}\selectfont \textBF{\textcolor{blue}{0.258$\pm$0.004}}} \\
    \arrayrulecolor{black}\bottomrule
    \end{tabular}
    \end{minipage}
\end{table}

To understand the impact of our design choices, we study the model performance under different ablations i.e.\ without the diagnosis encoder and LSTM components. 

Table~\ref{tab:ablation}a shows the performance of the LSTM-GNN models without a diagnosis encoder (denoted LSTM-GNN*). Firstly, we see that all of the LSTM-GNN* models easily outperform LSTM*, indicating that the patient graph alone can be an informative representation of diagnosis data.

When we consider the impact of the graph only vs.\ the combination approach (i.e.\ LSTM-GNN* vs.\ LSTM-GNN in Table~\ref{tab:results}), we see that the combined approach in LSTM-GNN produces the best results. However, the difference is more marginal for LOS, which suggests that the graph confers the larger benefit for LOS, whereas the encoder is more important for IHM. This can be explicitly verified by comparing the LSTM-GNN* to LSTM; where the LSTM-GNN* models do indeed outperform on LOS, but not IHM.

Table~\ref{tab:ablation}b shows the performance of the graph on the raw time series (i.e.\ no LSTM component). Both of these models perform significantly worse than their respective LSTM-GNNs, which validates the need for an LSTM for time series processing.

\begin{figure}[ht]
	\centering
	\includegraphics[width=0.73\textwidth]{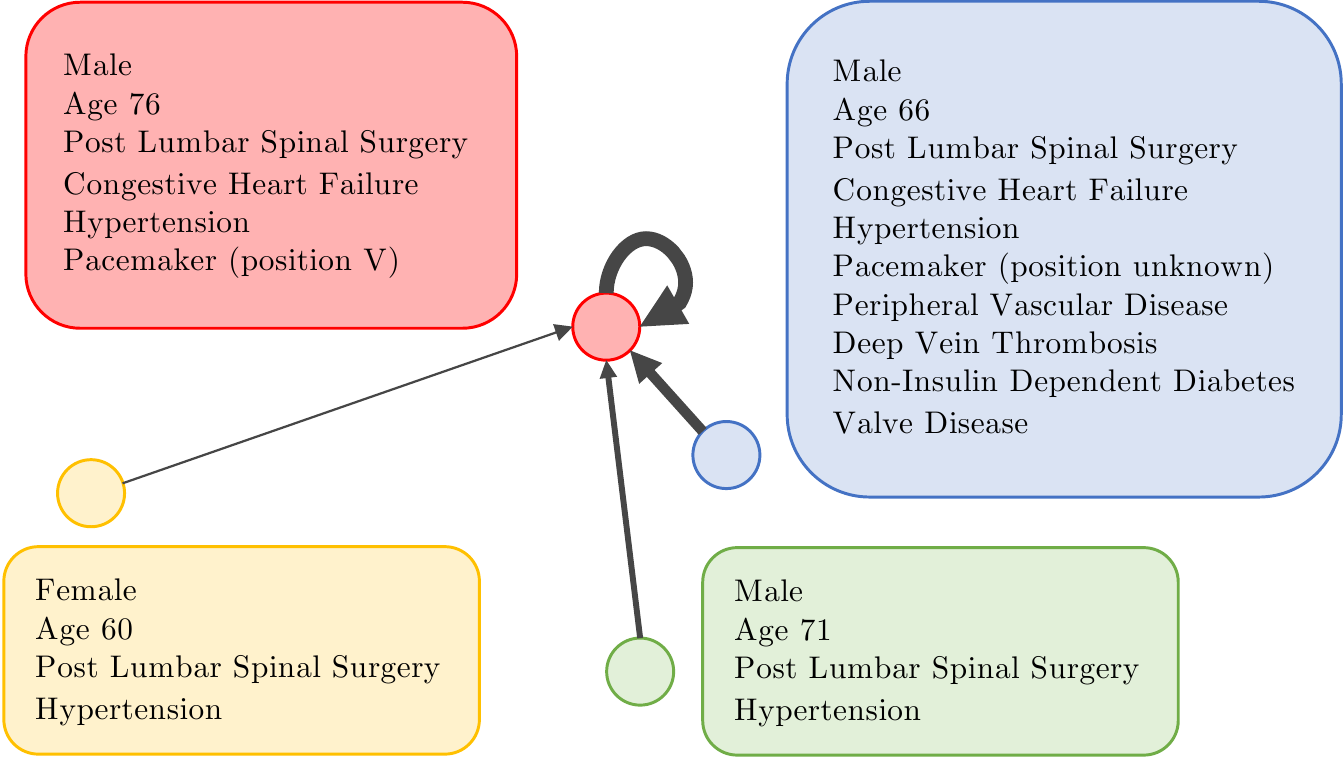}
	\caption{\small An example showing graph attention weights in LSTM-GAT* (indicated by edge thickness).}
	\label{fig:gat}
\end{figure}

\runinhead{Interpretability}
The LSTM-GAT model provides the additional benefit of assigning attention weights to their edges, meaning we can qualitatively assess what the model is examining. Figure~\ref{fig:gat} depicts a 76 year-old post-surgical patient and his neighbours. Typically, surgical patients have shorter stays \citep{gruenberg}, but this patient has congestive heart failure which is associated with high mortality and longer recovery times~\citep{heartfailure}. We see that a learned LSTM-GAT* behaves as we expect -- placing the highest weighting on the neighbour who shares his congestive heart failure diagnosis.

\section{Discussion}
In this work, we have proposed and evaluated a new LSTM-GNN architecture for diagnosis processing. Our results demonstrate that the representation of diagnoses as a graph confers an independent and substantial performance benefit when combined with the commonly applied encoder approach on the LOS task. This makes intuitive sense when considering the different architectures: The encoder method may be preferable for representing common comorbidities that confer strong correlations with the prediction task e.g. sepsis. However, the graph method provides a context for rarer patterns of disease by presenting examples cases in the local neighbourhood. Note that the graph may also help to augment the data where the quality in the original patient is poor. Since these approaches offer complementary insights, their respective contributions can be combined to obtain better performance in the LSTM-GNN.

\runinhead{Performance differences between LOS and IHM} The graph method is particularly strong on the LOS task, which may be attributable to the increased reliance on operational factors for LOS e.g.\ different discharging practices~\citep{Couturiere012287}, which in turn depend on the diagnoses. This is not upheld on IHM however, possibly because the vital signs and a few common diagnoses (which can be easily extracted from the diagnosis encoder) remain the most reliable predictors of mortality risk.

\runinhead{Future Work} A natural extension is to include shared medications and procedures in the similarity scores. We also plan to characterise the sensitivity of LSTM-GNN to parameters $\alpha$, $c$ and $k$. Nevertheless, our results thus far show that graph representation of sparse EHR data is a potentially rewarding avenue for future research.

\begin{acknowledgement}
This work was supported by the Engineering and Physical Sciences Research Council (EPSRC) under Grant No.: DTP (EP/R513295/1), MOA (EP/S001530/), Samsung AI, the Armstrong Fund, the Frank Edward Elmore Fund, and the School of Clinical Medicine at the University of Cambridge. Additionally we thank Cristian Bodnar, C\u{a}t\u{a}lina Cangea, Louis-Pascal Xhonneux, Stephanie Hyland, and anonymous reviewers at W3PHIAI-21 for their feedback.
\end{acknowledgement}

\bibliographystyle{spbasic}
\bibliography{references}

\section*{Appendix}
\subsection*{Dynamic LSTM-GNNs}

In our paper we propose a fixed patient graph constructed using diagnoses. However, here we investigate whether a useful graph can be learnt dynamically from the time series alone (in the absence of diagnoses). Inspired by Dynamic Graph CNNs \citep{dgcnn}, we explore a \emph{dynamic} variant of LSTM-GNN. Here we train an LSTM on the time series $x_{1:T}$ with mini-batching, each time computing the pairwise Euclidean distance of the hidden vectors $h_{T}$ in the batch. Again, we apply $k$-NN to obtain the graph.

\begin{table}[ht!]
  \caption{\small Performance of various dynamic LSTM-GNNs compared to LSTM*. These models do not have diagnoses in their static features; they create the graph from the temporal features alone.}
  \label{tab:resultsappendix2}
  \centering
  \scriptsize
  \begin{tabular}{p{1.6cm}p{1.2cm}p{1.17cm}p{1cm}p{0.87cm}p{0.87cm}p{1.2cm}p{1.2cm}p{1.2cm}}
    \toprule
        & \multicolumn{2}{c}{\scriptsize{\textBF{In-Hospital Mortality}}} & \multicolumn{6}{c}{\scriptsize{\textBF{Length of Stay}}} \\
        \scriptsize{\textBF{Model}} & \scriptsize{\textBF{AUROC}} & \scriptsize{\textBF{AUPRC}} & \scriptsize{\textBF{MAD}} & \scriptsize{\textBF{MAPE}} & \scriptsize{\textBF{MSE}} & \scriptsize{\textBF{MSLE}} & \scriptsize{\boldmath{$R^2$}} & \scriptsize{\textBF{Kappa}} \\
    \arrayrulecolor{black}\specialrule{0.7pt}{2pt}{3.5pt}
        \scriptsize{LSTM*} & {\fontsize{5.5}{5}\selectfont 0.837$\pm$0.001} & {\fontsize{5.5}{5}\selectfont \textBF{\textcolor{blue}{0.390$\pm$0.004}}} & {\fontsize{5.5}{5}\selectfont 1.97$\pm$0.01} & {\fontsize{5.5}{5}\selectfont \textBF{\textcolor{blue}{49.4$\pm$0.6}}} & {\fontsize{5.5}{5}\selectfont 17.6$\pm$0.2} & {\fontsize{5.5}{5}\selectfont 0.398$\pm$0.004} & {\fontsize{5.5}{5}\selectfont 0.089$\pm$0.008} & {\fontsize{5.5}{5}\selectfont 0.224$\pm$0.006} \\
        \scriptsize{Dyn. GCN*} & {\fontsize{5.5}{5}\selectfont \textBF{\textcolor{blue}{0.839$\pm$0.001}}}$^\ddag$ & {\fontsize{5.5}{5}\selectfont 0.388$\pm$0.002} & {\fontsize{5.5}{5}\selectfont \textBF{\textcolor{blue}{1.96$\pm$0.01}}}$^\ddag$ & {\fontsize{5.5}{5}\selectfont 50.2$\pm$1.0} & {\fontsize{5.5}{5}\selectfont \textBF{\textcolor{blue}{17.0$\pm$0.1}}}$^\ddag$ & {\fontsize{5.5}{5}\selectfont \textBF{\textcolor{blue}{0.387$\pm$0.002}}}$^\ddag$ & {\fontsize{5.5}{5}\selectfont \textBF{\textcolor{blue}{0.117$\pm$0.007}}}$^\ddag$ & {\fontsize{5.5}{5}\selectfont \textBF{\textcolor{blue}{0.251$\pm$0.005}}}$^\ddag$ \\
        \scriptsize{Dyn. GAT*} & {\fontsize{5.5}{5}\selectfont 0.832$\pm$0.001}$^\dag$ & {\fontsize{5.5}{5}\selectfont 0.358$\pm$0.005}$^\dag$ & {\fontsize{5.5}{5}\selectfont 1.97$\pm$0.01} & {\fontsize{5.5}{5}\selectfont 50.2$\pm$1.4} & {\fontsize{5.5}{5}\selectfont 17.3$\pm$0.1}$^\ddag$ & {\fontsize{5.5}{5}\selectfont 0.393$\pm$0.003}$^\ddag$ & {\fontsize{5.5}{5}\selectfont 0.105$\pm$0.007}$^\ddag$ & {\fontsize{5.5}{5}\selectfont 0.236$\pm$0.006}$^\ddag$ \\
        \scriptsize{Dyn. MPNN*} & {\fontsize{5.5}{5}\selectfont 0.837$\pm$0.001} & {\fontsize{5.5}{5}\selectfont 0.389$\pm$0.002} & {\fontsize{5.5}{5}\selectfont \textBF{\textcolor{blue}{1.96$\pm$0.01}}}$^\ddag$ & {\fontsize{5.5}{5}\selectfont 50.0$\pm$1.1} & {\fontsize{5.5}{5}\selectfont 17.1$\pm$0.1}$^\ddag$ & {\fontsize{5.5}{5}\selectfont 0.389$\pm$0.002}$^\ddag$ & {\fontsize{5.5}{5}\selectfont 0.113$\pm$0.007}$^\ddag$ & {\fontsize{5.5}{5}\selectfont 0.248$\pm$0.005}$^\ddag$ \\
        \bottomrule
    \end{tabular}
\end{table}

Table~\ref{tab:resultsappendix2} shows that LSTM* and dynamic LSTM-GNNs generally perform similarly on IHM, but the dynamic LSTM-GNNs have an advantage on the LOS task. We also observe that the dynamic LSTM-GCN* model, despite not having access to diagnoses, performs similarly to LSTM (second row of Table~\ref{tab:results}). This suggests that relating patients via a graph structure has value for modelling patient outcomes independently of diagnoses. This is possibly because where the data is poor quality or missing, the model can rely more on the neighbouring patients. However, the most visible gains (Table~\ref{tab:results}a) still come from using diagnoses for the graph construction. 

\subsection*{Implementation and Hyperparameter Search Methodology}
\label{hyperparamsearch}
For each model, we conducted 10 random hyperparameter trials. The hyperparameter search ranges and selected values can be found in our repository: \url{https://github.com/EmmaRocheteau/eICU-GNN-LSTM}.

All deep learning methods were implemented in PyTorch and optimised using Adam \citep{KingmaB14}. We used PyTorch Lightning~\citep{falcon2019pytorch} and Tune to structure our experiments and easily compare different hyperparameter choices. The maximum number of epochs was 25, although many models finished before this due to early stopping.

\end{document}